# A frame semantics based approach to comparative study of digitized corpus

Abdelaziz Lakhfif

LRSD Laboratory, Department of Computer Science,
University Ferhat Abbes Setif 1, Setif, Algeria
abdelaziz.lakhfif@univ-setif.dz

Mohamed Tayeb Laskri

LRI Laboratory, Department of Computer Science,
Badji Mokhtar University, Annaba, Algérie
laskri@univ-annaba.org

*Abstract*—in this paper, we present a corpus linguistics based approach applied to analyzing digitized classical multilingual novels and narrative texts, from a semantic point of view. Digitized novels such as "the hobbit (Tolkien J. R. R., 1937)" and "the hound of the Baskervilles (Doyle A. C. 1901-1902)", which were widely translated to dozens of languages, provide rich materials for analyzing languages differences from several perspectives and within a number of disciplines like linguistics, philosophy and cognitive science. Taking motion events conceptualization as a case study, this paper, focus on the morphologic, syntactic, and semantic annotation process of English-Arabic aligned corpus created from a digitized novels, in order to re-examine the linguistic encodings of motion events in English and Arabic in terms of Frame Semantics. The present study argues that differences in motion events conceptualization across languages can be described with frame structure and frame-to-frame relations.

*Keywords— digitized corpus, Arabic, motion events, lexicalization patterns, typology of languages, frame Semantics,*

## I. INTRODUCTION

Corpus linguistics tools and methods have been extensively used in digital humanities, last decades. From the beginning of computer history, corpus linguistics was used as an analysis tool and quantitative approaches for language-use statistics, such as word frequency computing, concordance, grammatical word frequency etc. Recently, studies within the fields of IA, cognitive psychology, computational linguistics, cognitive linguistics, etc. have focused on the power of narrative texts and novels and their translation into other languages, to bring evidences of the relation between thought and language. Levelt argued that, in language production, *grammatical encoding comprises both the selection of appropriate lexical concepts (entries in the speaker's vocabulary) and the assembly of a syntactic framework* [5]. It was argued that, at an early age, speaker of any language is trained to fit its native language typology's requirements by learning to determine which 'aspects of the mental image' *are realized in the form of grammatical marking in the native language* [6]. So, in order to encode event conceptual elements, speaker's process of producing language involves specific cognitive mechanism that fits thought components into most available linguistic forms provided by speaker's native language. Such a cognitive process of mapping thought to language was amply studied by researchers in crosslinguitic studies, especially the description of motion events, which has been, for a long time, a preferable area for contrastive research [3], [2], in particular the description of manner and path of motion which raise interesting issues of both typology and language use [8], [9], [10], [2]. Thus, cross-linguistic variation with respect to the semantic components distribution among motion events expressions have evoked for a long time controversial questions about the link between languages and thought [12], [18], thus posing serious challenges for machine translation systems, such as conflational divergences, which has been largely described and discussed in [14].

Two of the most important classical novels used for cross-lingual studies, by scholars in several desciplines [6], [4], [30], [31], were the chapter 14 of 'the hound of the baskervilles' and the chapter 6 of 'the hobbit'.

In this paper, we present our attempt to carry out a qualitative comparison of the distribution of semantic components and linguistic encodings of motion events expressions in English and Arabic in the light of Talmy's dichotomy typology (Talmy [8]) and Slobin's "thinking for speaking" hypothesis (Slobin [13]). Using our semi-automatic analysis and annotation tool to digitize and annotate those chapters, with morpho-lexical, syntactic and semantic roles we were able to compare different conceptualization of the same event in different languages.

Hence this article aims to use corpus linguistics technics and tools applied to digitized calassical narrative texts and novels, containing motion events expressions, to reassess two influential theories about "thought and language"; that is, Talmy's typology of langage and Slobin's 'thinking for speaking' hypothesis, from Fillmore's Frame Semantics theory point of view .

The rest of the paper is structured as follow: Section 2 gives an introduction to the digitization and annotation process of narrative texts. Section 3 gives an overview on the theoretical background of our cognitive and computational based analysis of motion events, by introducing Talmy's dichotomy typology, Slobin's "thinking for speaking" hypothesis and the Fillmore's Frame Semantics theory. In section 4, a Frame based crosslinguitic analysis of motion event is presented and we conclude in section 5.

## II. WHY DIGITIZED NOVELS?

Following Talmy in his influential works on language typology, and from "*frog where are you*" to the fanciful story of "*the Hobbit*", Slobin and his colleagues have established a set of criteria that may help in the classification of languages studies with regard to the lexicalization patterns of motion events. Cross-linguistics comparison of motion events encoding can be carried out in terms of two vectors of factors: the degree of manner of motion salience and the degree of path of motion elaborations. As regard the degree of manner salience, in addition to the tendency to encode manner in the main verb slot, the manner salience can be also evaluated based on the richness of the language's lexicon of manner verbs, the language capability to describe directional motion with manner verbs of descent and the degree of use of manner verbs with grounds etc. As regard the degree of path of motion elaborations, in addition to the tendency to encode path information in the verb root and expressing manner of motion via adverbials, languages can be compared with regard to the canonical segmentation of paths as well as the relative ease of building complex-path constructions. In an attempt to assess and validate Talmy's and Slobin's hypothesis, we have built a multilingual aligned parallel corpus from digitized narrative texts selected from several famous novels in the last century. As a first step, we focus, on the chapter 6 of the hobbit.

### A. Text digitization, collection and annotation

Our digitized narrative corpora collection provides original digitized texts from the selected classical novels in their original version and their translation supported by three (03) languages (English, French and Arabic) (Fig. 2, Fig. 3). To this end, and in order to carry out a computational readable text digitization of multilingual narrative texts and novels, we have developed a java based tool which can deal with different languages specificities (Fig. 2).

## Out of the Frying-Pan into the Fire

Bilbo had escaped the goblins, but he did not know where he was. He had lost hood, cloak, food, pony, his buttons and his friends. He wandered on and on, till the sun began to sink westwards – *behind the mountains*. Their shadows fell across Bilbo's path, and he looked back. Then he looked forward and could see before him only ridges and slopes falling towards lowlands and plains glimpsed occasionally between the trees.

**Fig. 1.** A snapshot from the Tolkien's book 'the hobbit'

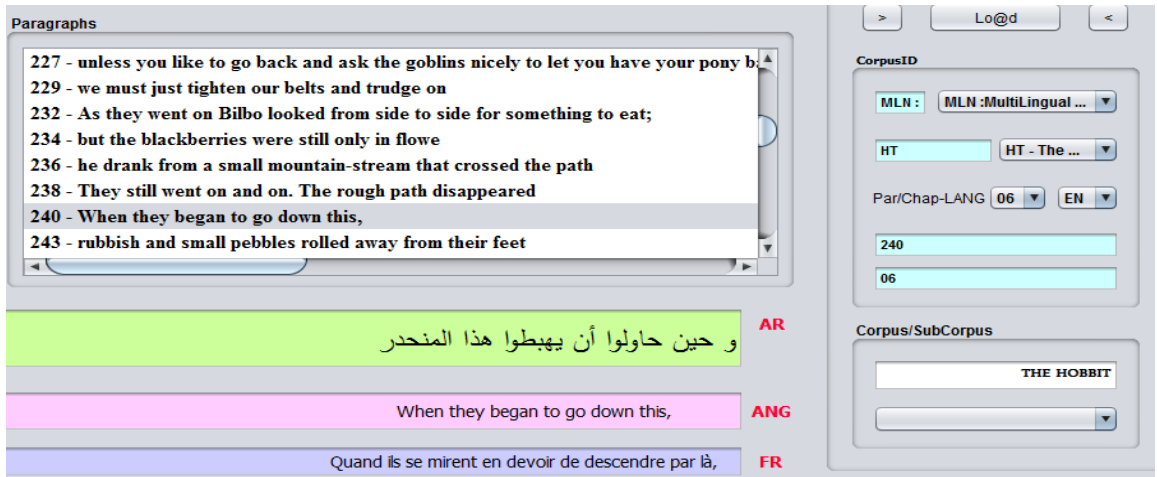

**Fig. 2.** Our java based tool for text digitization

As a first step, we have digitized the original version of the chapter 6 of the hobbit and the chapter 14 of the hound of the Baskervilles and theirs French and Arabic translations.

Collected data is stored in XML format and organized by novel, chapter and paragraph to form a parallel aligned corpus (Fig. 3).

```
<prg pID="p70">
    <p lang="AR" ID="239">و حين حاولوا هذا المنحدر/p>
    <p lang="EN" ID="240">When they began to go down this,</p>
    <p lang="FR" ID="241">Quand ils se mirent en devoir de descendre par là</p>
</prg>
<prg pID="p71">
    <p lang="AR" ID="242">تدحرجت القاذورات و الحصى الصغير من بين أقدامهم/p>
    <p lang="EN" ID="243">rubbish and small pebbles rolled away from their feet</p>
</prg>
```

**Fig. 3.** An XML based transcription of the digitized text

## B. Description of the Annotation Schemes

To perform a fine grained analysis and characterize the encoding system of path and manner in Arabic, we use our multi-level annotation tool [29] to annotate aligned texts with morphologic, syntactic (constituent and dependency-based syntactic), and semantic levels of description. The tool used for Arabic text annotation is compatible with FrameNet design and annotation schemes, providing an XML-data transcription (Fig. 4). Our annotation tool uses both AraMorph, Buckwalter Arabic Morphological Analyzer (BAMA) [15] for lexico-morphological analysis and Arabic WordNet (AWN) [17] for augmenting BAMA lexicon with semantic features. AWN-based analysis improves the token with lexical knowledge, which has proved helpful to bridge the semantic gaps and the coverage issue using its rich network of semantic relations between words. Also, AWN provides, for each analysed word, its equivalent ontological class (SUMO). For example, the syntactic dependency analysis output encodes the sentence structure as an XML tree representation in which, the predicate is the root (node 0), nouns can be head nodes which are linked to the predicate by syntactic labels (subject, object .Etc.), and then considered as predicate-arguments. Otherwise, noun sub-categories such as adjective, circumstantial accusative, accusative cognate etc. can be modifiers of either a head noun, or modifiers of verbs.

**Fig. 4.** A fragment of automatic analysis and multi layers annotation of an Arabic sentence

In addition to Arabic text annotation, the tool performs also the task of semantic roles annotation with triple information (FE/GF/PT) of English text (Fig. 5). for the annotation procedure, firstly we have aligned the pairs of English motion events expressions occurring in the Chapter 6 of the hobbit based on its original sentence and paragraph number. Secondly, we annotated the same sentences in Arabic and in its English counterpart, usually with several levels of description, (lexico-morphological, syntactic and semantic relevant information). Finally, we are investigating the semantics of lexical units (the predicate of the motion event within its conceptual structure class) and the grammatical structures in each language. The three levels of annotation on the multilingual aligned parallel corpus will allow us to easily examine the different lexicalization patterns of a given scene. We have applied a FrameNet-Like analysis to the original given chapter of the Hobbit along with its translation equivalent in Arabic.

Clear Sentences   Turn Colors Off

[X] CNI، بل إنْتَظَر حَتَّى نَجَح الهوبيتُ في الخُرُوج فَصَعُوبَةٌ بالِغَةٍ ثُمَّ قَفَزَ الْقَزَمُ إِلى الفُرُوعِ هُوَ أَيْضاً في الوَقْتِ المُناسِبِ تَماماً [X]

[X] He waited till he had CLAMBERED off his shoulders into the branches, and then he jumped for the branches himself. Only just in time!

[X] نَهَشَ أَحَدُ الذِئابِ عَباءَةِ دوري بِأَناهِ ثُمَّ تَأَرْجَحَ حَتَّى كادَ يَنالُ مِنهُ [X]

[X] A wolf snapped- at his cloak as he SWUNG up, and nearly got him.

[X] و خِلال دَقيقَةٍ واحِدَةٍ تَجَمَّعَت مَجْمُوعَةٌ كَبيرَةٌ مِن الذِئابِ تُعوي حَوْلَ الشَجَرَةِ مِن كُلِّ جانِبٍ وَ تَقْفِزُ خَلْفَ جُذُوعِها باعُيُنٍ بَرّاقَةٍ وَ أَلسِنَةٍ مُتَذَلِيَةٍ وَ أَشداقٍ يَسيلُ مِنها اللُعابُ DNI [X]

[X] In a minute there was a whole pack of them yelping all round the tree and LEAPING up at the trunk with eyes blazing and tongues hanging out.

**Fig. 5.** Annotation of bilingual text using our Annotation Tool.

## III. THEORETICAL BACKGROUND

In his influential work on typological classification of grammatical constructions of complex events, Talmy [13], has proposed that languages can be classified into two main typological categories on the basis of the way they map semantic components of motion events expressions onto linguistic forms, especially, where the path information, as a "core schema", is encoded, inside or outside the verb. Subsequently, Talmy's two-category typology has lead to a big and open workshop on the encoding system of motion events, from a crosslinguistic perspective. In accordance with Talmy's hypothesis, it is likely that English [18], Russian [20] and German [22], among other languages, are relatively more concerned with the manner of motion and thus they belong to satellite-framed languages. Oppositely, Spanish [20], French [22], Japanese [10], [24], [2], Basque [26], Italian [20] and Semitic languages [18], [20], characteristically, conflate motion with the core information of path movement and, thus, they are categorized along with verb-framed languages.

### A. Interacting factors in encoding system of motion events

in this paper, we extend the investigation bases with a conceptual and computational cognitive approach, using FrameNet principles for the encoding of the components of distributed semantics conveyed by motion events expressions in a representative aligned bilingual corpus. We built on previous Frame Semantics based contrastive studies [30], [31], [3] in which authors have pointed out that analyzing motion events in terms of Frame Semantics principles throw a new light on several unexplained regularities of translation divergences. More interestingly, they argued that differences in motion events conceptualization across languages can be described in terms of frame structure and frame-to-frame relations. Here, we review these hypotheses and their implications, in particular regarding Arabic language, and present new empirical evidences by exploring the consequences of the differences in lexicalization patterns in frame structures and how motion events elements are characterized in terms of frame participants (*Frame Elements*) and syntactic realization within the Frame Semantics theory.

### B. Frame Semantics principles

Frame Semantics [35], [33] is a research program in empirical semantics for describing word meanings and grammatical constructions in terms of underlying conceptual structures. Frame Semantics idea rests on Fillmore's uncontroversial assumption that in order to understand the meanings of the words in a language we must first have knowledge of the conceptual structures, or semantic frames, which provide the background and motivation for their existence in the language and for their use in discourse [34]. From this point of view, A frame is *any system of concepts related in such a way that to understand any one of them you have to understand the whole structure in which it fits; when one of the things in such a structure is introduced into a text, or into a conversation, all of the others are* automatically made available [16]. A Frame (Fig. 6) is a cognitive structure that depicts schematic representations of situations involving various participants, props, and other conceptual roles, each of which is a frame element (FE). Such a structure can be seen as "a script-like conceptual structure that describes a particular type of situation, object or event". Semantic participants or frame elements (FEs) represent required semantic components that play central roles in the event or inherent parts of the object.

Unlike traditional thematic roles or case roles [21], FEs are frame-specific role names and they have a semantic types (STs). Semantic types (STs) are, also, arranged in ontological-like hierarchy. Some FEs are necessary (core) to understand the message conveyed in the sentence. One of the most important characteristics of Frame Semantics theory is its rich set of relationships between Frames, such as *inheritance, precedes, has_subframe, causative_of* and *inchoative_of* relations, which play key roles in translation-divergence handling. In Frame Semantics tenet, a set of semantically related words are understood with a particular Frame as background, and thus they are grouped as frame-bearing lexical units (LUs). For instance, the verbs *angle, descend, dip, drop, fall, plunge, plummet, rise, slant* and *topple* belong to the *Motion_directional* Frame, as they all have as background a situation in which a *Theme* moves in a certain *Direction* which is often determined by gravity or other natural, physical forces. As a result, Berkeley FrameNet project (BFN) is an online lexical database [23] that implement the idea of Frame Semantics theory, supported by corpus evidence. In Frame Semantics, a word represents a category of experience [4]. In sum, a linguistic expression is understood with regard to a set of background knowledge which are associated with lexical units and the grammatical construction of the linguistic expression. FrameNet approaches the motion events by a general *Motion* Frame, a cogitive structure, involving little more than location changes whose components are a starting point (*Source*), a trajectory (*Path*), and a destination (*Goal*).

*Frame* inheritance and *Frame-to-Frame* relations elaborate fine grained notion of motion having regard to the *Source*, *Goal* or *Path* are profiled, and thus motion verbs are organized according to their shared backgrounds.

Frames like *Arriving*, *Departing*, *Traversing* and *Motion_directional* are elaborations of this general Frame.

| MOTION_DIRECTIONAL FRAME | |
|---|---|
| Definition | In this frame a **Theme** moves in a certain **Direction** which is often determined by gravity or other natural, physical forces. The **Theme** is not necessarily a self-mover. *The paper fell to the floor.* *The girl dropped 13 stories to her death.* |
| Core FEs | 1. **Area** [Area] 2. **Direction** [dir] **Excludes:** Area 3. **Goal** [Goal] **Semantic Type:** Goal **Excludes:** Area 4. **Source** [Src] **Semantic Type:** Source **Excludes:** Area 5. **Path** [Path] **Excludes:** Area 6. **Theme** [Thm] **Semantic Type:** Physical_object |
| frame-frame relations | Inherits from: *Motion* Is Used by: *Sidereal_appearance* |
| English LUs | angle.v, descend.v, dip.v, drop.v, fall.v, plunge.v, plummet.v, rise.v, slant.v, topple.v |
| Arabic LUs | اِنْحَدَرَ.v ، اِنْحَرَفَ.v ، اِنْحَنَى.v ، اِنْخَفَضَ.v ، اِنْهَارَ.v ، تَدَاعَى.v ، تَدَحْرُج.n ، تَدَخْرَج.v ، تَصَاعَد.v ، تَعَالَى.v ، سَقَطَ.v ، مَالَ.v ، نَزَلَ.v ، هَبَطَ.v ، صُعُود.n |

Fig. 6.    A lexical Frame Description

## IV. A CASE STUDY : FRAME SEMANTICS CONCEPTUALIZATION OF MOTION EVENTS

During the last decade, cross-linguistic text analysis has received an increasing interest in the FrameNet community, with special focus on the contrastive analysis of motion events description within the frame semantics approach.

Motion events conceptualization has received growing interest. In particular, the interface between linguistic forms and their corresponding semantic components by allowing for the investigation of similarities and differences in how different languages express similar meanings. So, following attempts for other languages such as German, Spanish, Japanese and Hebrew [30], we extend our previous contrastive analysis of motion events expressions in Arabic and English with an analysis based upon use of Frame Semantics theory. That is, we carry out a frame based analysis of the same English text and its Arabic translation "Fig. 5". Ohara [31] claims that Frames conceptual structure, frame-to-frame relations and the FrameNet methodology allow us to compare languages at a more detailed level than previous studies. In the light of Frame Semantics theory, Ellsworth et al. [30] presented the first contrastive analysis. They have compared the frame-evoking predicates in different languages. The work reveals details about types of translation shift and contrastive aspects that is not covered by Talmy's and Slobin's studies [31], such as focusing on action vs. focusing on state etc. In addition, Ohara [31] argues that understanding how languages characterize the same scene differently involves taking into account the interaction between grammar construction and frame-evoking predicates [31].

TABLE I.    TYPES OF FRAMES CHARACTERIZING MOTION EVENTS IN ENGLISH AND ITS ARABIC COUNTERPART [32].

| Evoked Frames (Ehglish version) | Evoked Frames (Arabic translation) | # of expressions |
|---|---|---|
| Self_motion | Self_motion | 56 |
| Self_motion | Motion_directional | 01 (نَزَلَ) |
| Self_motion | Arriving | 02 (عاذ – إقترب) |
| Self_motion | Manipulation | 01 (تَعلَّق) |
| Motion | Motion_directional | 02 (تَدَحْرَج) |
| Motion | Self_motion | 02 (إنْزَلَق) |
| Motion_directional | Motion_directional | 04 (سَقَط، وَقَع) |
| Motion_directional | Cause_motion | 01 (أوْقَع) |
| Cause_to_move_in_place | Manipulation | 01 (تَعلَّق) |
| Fleeing | Fleeing | 01 (هَرَب) |
| Dispersal | Self_motion | 01 (تَفَرَّق) |

In order to examine whether English and Arabic languages conceptualize motion events in the same or different Frame Semantics terms, we have elaborated a comparative table (Table I.) of evoked Frames of motion events of human-like creatures in English original of the hobbit and its translation equivalent in Arabic [32]. As a result, the table 1, elaborated from annotated aligned sentences (Fig.5), shows the divergence as regard the conceptual schemas (Frames) evoked by English and Arabic to characterize motion events. From 72 evoked Frames in the English original text, the Arabic translation evoked the same Frames for 61 sentences, characterizing thus a parallel conceptualization in 85% of the original English text. Despite the differences in language typology between the two languages (English/Arabic), this result appears to support the assumptions that the semantic Frames are language-independent cognitive structures.

## V. CONCLUSION

In this article, we have presented a corpus linguistics technics and tools applied to digitized calassical narrative texts and novels containing motion events expressions, to reassess two influential theories about "thought and language"; that is, Talmy's typology of language and Slobin' 'thinking for speaking' hypothesis. The assessing approach presented here built on Fillmore's Frame Semantics theory point of view. We have examined the characterization of motion events expressions in English and Arabic based upon use of FrameNet principles for the coding of the components of distributed semantics, conveyed by motion events expressions.

An important finding of our contrastive study is that, despite the fact that Arabic and English are typologically opposite, almost 90% of English motion events in "the hobbit" are characterized by the same semantic Frames in the Arabic translation, providing thus a high degree of framing parallelism.

In the futur works, we plan to extend our investigation to include many more multilingual stories and narrative texts.

## References

[1] M. Minsky,"A framework for representing knowledge", In P. Winston (Ed.), *The Psychology of Computer Vision*, McGraw-Hill, 1975.

[2] J. Beavers, B. Levin and S. W. Tham, "A morphosyntactic basis for variation in the encoding of motion events". *Journal of Linguistics*, *46*, 331-377, 2010.

[3] M.R.L Petruck, "Frame semantics", In Verschueren, J., Östman, J-O., Blommaert, J. & Bulcaen, C., eds. *Handbook of pragmatics*. Amsterdam: John Benjamins, p.1-13, 1996.

[4] M.R.L. Petruck, "Framing motion in Hebrew and English", 2008.

[5] K. Bock, and W., Levelt, "Language production" *Psycholinguistics: Critical concepts in psychology*, vol. *5*, p.405. 2002.

[6] D. I. Slobin, "Learning to think for speaking: Native language", *cognition, and rhetorical style. Pragmatics*, vol *1*(1), pp. 7-25, 1991.

[7] J. R. R. Tolkien, "The Hobbit or there and back again", *London: George Allen & Unwin, 1937*. [1- Trad.Arabe. H. Fahmy, M. Ghanim, 2008. " ( عودة و ذهايا أو) الهوبيت". Dar Lila-Boeken.] [2- Arabic transl. M. El Douakhly.. "الهوبيت".] [French transl. F. Ledoux, 1980. Bilbo le Hobbit. Paris: Hachette.]

[8] L.Talmy, "Path to realization: A typology of event conflation". In *Annual Meeting of the Berkeley Linguistics Society* , Vol. 17, No. 1, pp. 480-519, 1991.

[9] R., Berman and D. Slobin, "Relating events in narrative: A crosslinguistic developmental study". *Hillsdale, NJ: Erlbaum*, 1994.

[10] Y. Matsumoto, "Typologies of lexicalization patterns and event integration: Clarifications and reformulations". *Empirical and theoretical investigations into language: a festschrift for Masaru Kajita*, pp. 403-418, 2003.

[11] E. Goffman, "Frame Analysis", New York. Harper & Row, 1974.

[12] B.L. Whorf, "Language, thought, and reality: selected writings of….", Edited by John B. Carroll. 1956.

[13] D. I. Slobin, "From 'thought and language' to 'thinking for speaking'". *In J. Gumperz, & S. Levinson (Eds.), Rethinking linguistic relativity*, pp. 70–96, New York: Cambridge University Press, 1996.

[14] B.J. Dorr, "Machine translation divergences: A formal description and proposed solution", *Computational Linguistics, vol, 20(4),*1994.

[15] T. Buckwalter, "Buckwalter Arabic Morphological Analyzer 1.0". *Linguistic Data Consortium*, 2002.

[16] C.J., Fillmore, and C.F. Baker, "A frames approach to semantic analysis". In *B. Heine & H. Narrog (Eds.), The Oxford handbook of linguistic analysis*, pp. 313–340. Oxford: Oxford University Press, 2010.

[17] S. Elkateb, W. Black, H. Rodriguez, M. Alkhalifa, P. Vossen, A.Pease, and C.Fellbaum, "Building a WordNet for Arabic", *in Proceedings of The Fifth International Conference on Language Resources and Evaluation*, LREC, 2006.

[18] L, Talmy,. "Lexicalization patterns". *In Timothy A. Shopen (ed.) Language Typology and Syntactic Description* Vol. 3, pp. 57-149. Cambridge: Cambridge University Press, 1985.

[19] G. Bernini, "Word classes and the coding of spatial relations in motion events: A contrastive typological approach", *in Marotta, G., Lenci, A., Meini, L. and Rovai, F. (eds.), Space in language*: Proceedings of the Pisa International Conference, Pisa, ETS: pp.29-52, 2010.

[20] D.I. Slobin, "Two Ways to Travel: Verbs of Motion in English and Spanish". In: Shibatani, M. & S.A. Thompson (eds.) *Grammatical Constructions: Their Form and Meaning*. Oxford: Oxford University Press, pp 195-217. 1996.

[21] C. J. Fillmore, "The case for case", *In E. Bach & R. T. Harms (Eds.), Universals in linguistictheory*. New York: Holt, Rinehart & Winston,1968.

[22] R. Berthele, "The typology of motion and posture verbs: A variationist account". *Trends In Linguistics Studies And Monographs*, *153*, pp. 93-126, 2004.

[23] C.F. Baker, C.J. Fillmore and B. Cronin, "The structure of the FrameNet database", *International Journal of Lexicography*, vol *16*(3), pp. 281-296, 2003.

[24] Y.Sugiyama, "Not all verb-framed languages are created equal: The case of Japanese". In *Annual Meeting of the Berkeley Linguistics Society* , Vol. 31, No. 1, pp. 299-310, 2005.

[25] A. Lakhfif, and M.T. Laskri, "A frame-based approach for capturing semantics from Arabic text for text-to-sign language MT". *International Journal of Speech Technology*, pp. 1-26, 2015, doi: 10.1007/s10772-015-9290-8.

[26] I. Ibarretxe-Antuñano, "Path Salience in Motion Events" In: Jiansheng Guo et al. (eds.) *Cross-Linguistic Approaches to the Psychology of Language: Research in the Tradition of Dan Isaac Slobin*. NY: Psychology Press, pp. 403-414, 2009.

[27] R.W. Langacker, "Foundation of Cognitive Grammar", Vol 1. *Theoretical prerequisites*, Stanford. CA : Stanford University Press, 1987.

[28] J. M. Gawron, "Frame Semantic", 2008.

[29] A. Lakhfif, M.T. Laskri, and E. Atwell, "Multi-Level Analysis and Annotation of Arabic Corpora for Text-to-Sign Language MT", *in Proceedings of WACL'2 Second Workshop on Arabic Corpus Linguistics*, Lancaster, UK. pp.49-52, 2013.

[30] M. Ellsworth, K. Ohara, K. Subirats and T. Schmidt, "Frame-semantic analysis of motion scenarios in English, German, Spanish, and Japanese", *Paper presented at The 4th International Conference on Construction Grammar*, Tokyo, 2006.

[31] K.H., Ohara, " Frame Semantics in Action: A Frame -based Contrastive Text Analysis Using FrameNet", *In the 10th International Cognitive Linguistics Conference* , Poland, 2007.

[32] A. Lakhfif, M.T. Laskri, "L'analyse et l'annotation à base de FrameNet : contribution à l'étude contrastive des événements de mouvement en arabe et en anglais"**,** *Revue TAL-Traitement Automatique des Langues, vol. 58 − n° 3, pp. 67-95,2017*

[33] C.J. Fillmore, C. R. Johnson, and M.R.L. Petruck, "Background to FrameNet", *International Journal of Lexicography*, vol. 16, pp. 235–250, 2003

[34] C.J. Fillmore, "Frames and the semantics of understanding", *Quaderni di semantica*, vol. 6(2), pp.222-254, 1985.

[35] C.J. Fillmore, "Frame semantics", *Cognitive linguistics: Basic readings*, pp.373-400, 1982.